\documentclass[10pt,twocolumn,letterpaper]{article}

\usepackage{cvpr}
\usepackage{times}
\usepackage{epsfig}
\usepackage{graphicx}
\usepackage{amsmath}
\usepackage{amssymb}
\usepackage{booktabs}

\usepackage{makecell}
\usepackage{subcaption}

\usepackage[pagebackref=true,colorlinks]{hyperref}

\cvprfinalcopy 

\ifcvprfinal\fi
\begin{document}

\title{MeshGAN: Non-linear 3D Morphable Models of Faces}

\author{
\parbox{16cm}{\centering
    {\large Shiyang Cheng$^1$, Michael Bronstein$^1$, Yuxiang Zhou$^1$, Irene Kotsia$^2$, \\Maja Pantic$^1$, Stefanos Zafeiriou$^1$}\\
    {$^1$Imperial College London 
     $^2$Middlesex University London}}
}
\maketitle

\begin{abstract}
Generative Adversarial Networks (GANs) are currently the method of choice for generating visual data. Certain GAN architectures and training methods have demonstrated exceptional performance in generating realistic synthetic images (in particular, of human faces). 
However, for 3D object, GANs still fall short of the success they have had with images. 
One of the reasons is due to the fact that so far GANs have been applied as 3D convolutional architectures to discrete volumetric representations of 3D objects. 
In this paper, we propose the first intrinsic GANs architecture operating directly on 3D meshes (named as MeshGAN). Both quantitative and qualitative results are provided to show that MeshGAN can be used to generate high-fidelity 3D face with rich identities and expressions.
\end{abstract}

\section{Introduction}
\begin{figure}[ht!]
\begin{subfigure}[t]{0.95\linewidth}
    \begin{flushleft}
        \hspace{10pt} CoMA \hspace{6pt} MeshGAN
    \end{flushleft}
    \centering
    \vspace{-9pt}
    \includegraphics[width=0.95\linewidth]{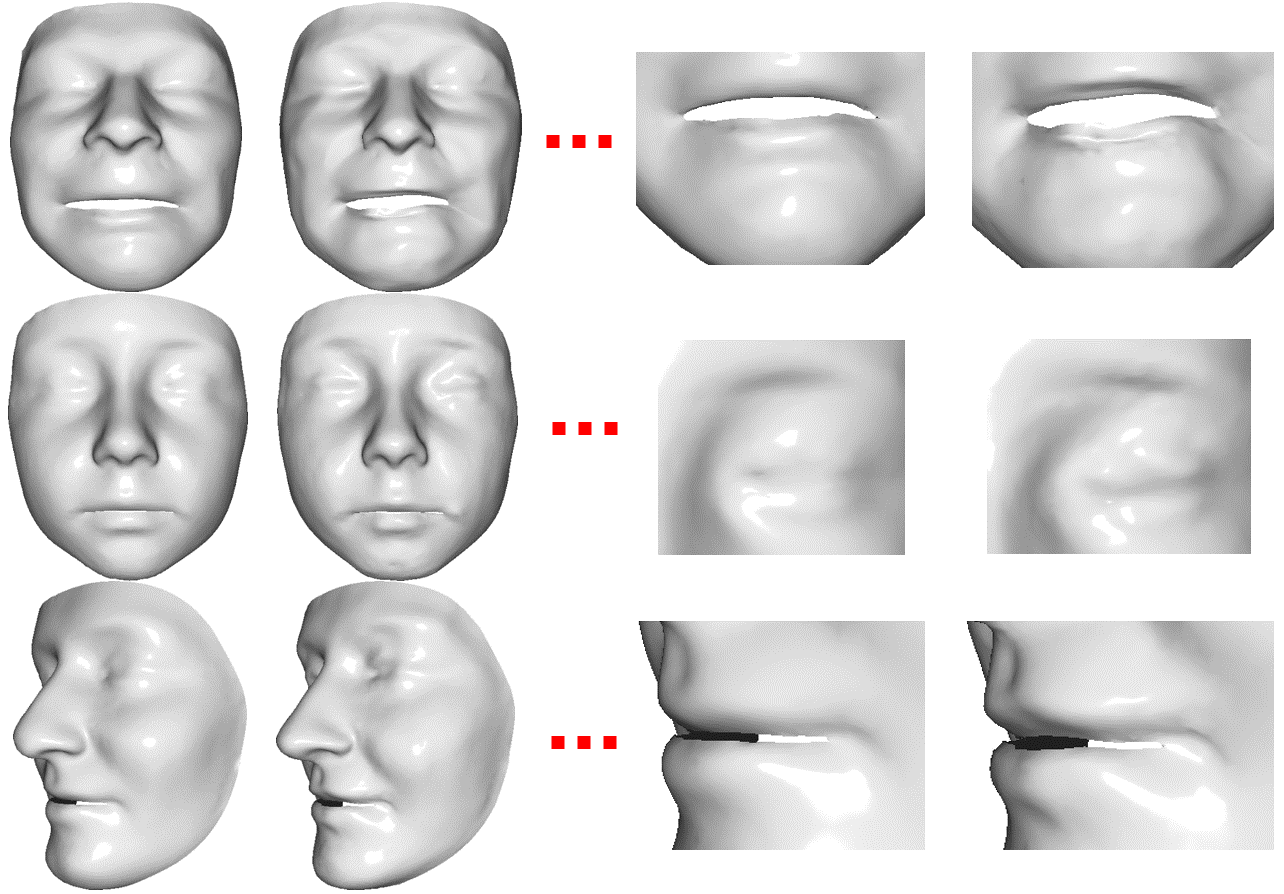}\\
    \vspace{-5pt}
    \caption{Exemplar reconstruction results of MeshGAN.}
\end{subfigure} \\
\begin{subfigure}[t]{0.95\linewidth}
    \centering
    \includegraphics[width=0.95\linewidth]{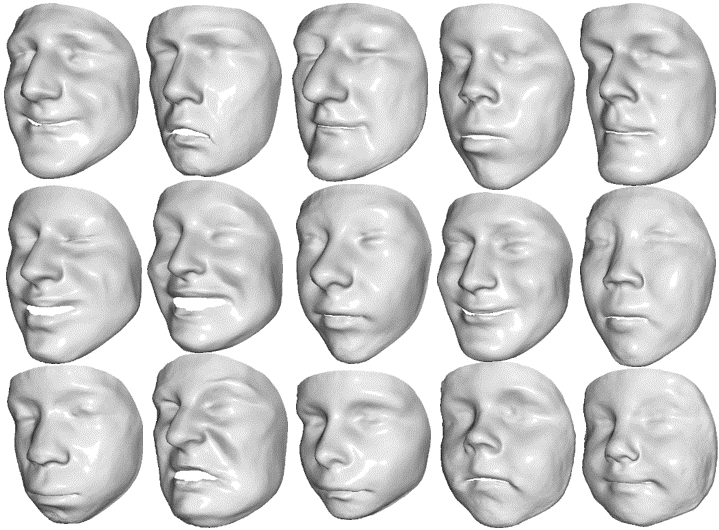}\\
    \vspace{-5pt}
    \caption{MeshGAN generation of identities and expressions.}
\end{subfigure}
\vspace{-8pt}
\caption{Qualitative reconstruction and generation of the proposed MeshGAN. Please zoom in to see more details.}
\label{fig:intro}
\end{figure}
Over the past few years deep Convolutional Neural Networks (CNNs) have emerged as the method of choice for the majority of computer vision tasks that require learning from data \cite{lecun2015deep, chen2018deeplab, he2016deep}. While the initial use of CNNs was mainly limited to classification/segmentation tasks \cite{chen2018deeplab,he2016deep}, the introduction of Generative Adversarial Networks (GANs) \cite{goodfellow2014generative} has expanded the application of deep convolutional architectures to image generation \cite{karras2017progressive,arjovsky2017wasserstein,berthelot2017began,radford2015unsupervised} and image-to-image translation and completion \cite{isola2017image,zhu2017unpaired,choi2017stargan}. Recently, strikingly realistic results have been shown by Nvidia using progressive GANs~\cite{karras2017progressive}.

Given the success of generative models in images, certainly there is a keen interest in replicating them for geometric data. 
In order to make convolutions/de-convolutions feasible, the current generative approaches still rely on crude shape approximations. For example, recent approaches either use discrete volumetric representations for the 3D shapes, which result in very low-quality shapes \cite{wu2016learning}, or they apply 1D convolutions combined with fully connected layers \cite{achlioptas2018learning} which do not take into account the local structure of 3D shapes.

Recently, the field of geometric deep learning on non-Euclidean (graph- and manifold) structured data has gained popularity \cite{bronstein2017geometric}, with numerous works on generalizing convolutional architectures directly on meshes. Intrinsic generative models are currently a key open question in geometric deep learning.   
Intrinsic auto-encoder architectures have recently been proposed for human body~\cite{litany2017deformable} and face~\cite{ranjan2018generating} meshes. Nevertheless, due to the lack of appropriate adversarial training, these auto-encoders retain only the low-pass shape information and lose most of the details. Furthermore, contrary to GANs, they do not offer a principled sampling strategy.
As of today, we are not aware of any successful intrinsic GAN for 3D mesh generation.


In this paper, we try to bridge this gap with the following contributions: 
\begin{itemize}
    \item We present the first intrinsic GANs architecture to generate 3D meshes using convolutions directly on meshes. Compared to approaches based on volumetric \cite{wu2016learning,achlioptas2018learning} or point cloud representations, our MeshGAN is able to generate meshes with high level of details.
    
    \item We present the first GAN architecture for 3D face generation. Contrary to the auto-encoder recently proposed in \cite{ranjan2018generating} that learns latent spaces where identity and expression are mixed, we can generate expression for arbitrary identities. 
    
    \item We conduct quantitative and qualitative experiments to verify the efficacy and effectiveness of MeshGAN on large scale 3D facial data. 
\end{itemize}
\section{Related Work}

\subsection{Geometric deep learning}

Geometric Deep Learning (GDL) is an emerging field in machine learning attempting to generalize modern deep learning architectures (such as convolutional neural networks) and the underpinning mathematical principles to non-Euclidean domains such as graphs and manifolds (for a comprehensive survey, the reader is referred to the recent review papers  \cite{bronstein2017geometric,hamilton2017representation,battaglia2018relational}). 
%
%

First formulations of neural networks on graphs 
\cite{gori2005new,scarselli2009graph} preceding the recent renaissance of deep learning, constructed learnable information diffusion processes. This approach has more recently been reformulated using modern tools such as gated recurrent units \cite{li2016gated} and neural message passing \cite{gilmer2017neural}.  
%
%
%
Bruna \etal \cite{bruna2013spectral,henaff2015deep} proposed formulating convolution-like operations in the spectral domain defined by the eigenvectors of the  Laplacian graph. One of the key drawbacks of this approach leading to high computational complexity is the necessity to explicitly perform the Laplacian eigendecomposition. 
However, if the spectral filter function can be expressed in terms of simple operations (scalar- and matrix multiplications, additions, and inversions), it can be applied directly to the Laplacian avoiding its explicit eigendecomposition altogether. Notable instances of this approach include ChebNets \cite{defferrard2016convolutional,kipf} (using polynomial functions) and CayleyNets \cite{levie2017cayleynets} (using rational functions); it is possible to generalize these methods to multiple graphs \cite{monti2017geometric} and directed motif-based graph Laplacians \cite{monti2018motifnet} using multivariate polynomials. 

Another class of graph CNNs are spatial methods, operating on local neighborhoods on the domain~\cite{duvenaud2015convolutional,monti2016geometric,atwood2016diffusion,hamilton2017inductive,velivckovic2017graph}. 
For meshes, the first such architecture (GCNN) used local geodesically polar charts \cite{masci2015geodesic}; 
alternative constructions were proposed using anisotropic diffusion (ACNN) \cite{boscaini2016learning} and learnable Gaussian kernels (MoNet) \cite{monti2016geometric}. SplineCNN \cite{fey2018splinecnn} uses B-spline kernels instead of Gaussians, offering significant speed advantage. FeastNet \cite{verma2018feastnet} uses an attention-like soft-assignment mechanism to establish the correspondence between the patch and the filter. Finally, \cite{lim2018simple} proposed constructing patch operators using spiral ordering of neighbor pixels. 

The majority of the aforementioned works focus on extracting features on non-Euclidean data (\eg, graphs, meshes, and etc.) for classification purposes and limited work has been done towards training generative models. One of the fundamental differences between classical Euclidean generative models (such as auto-encoders \cite{kingma2013auto} or Generative Adversarial Networks (GANs) \cite{goodfellow2014generative}) is the lack of canonical order between the input and the output graph, thus introducing some kind of graph correspondence problem to be solved. In this paper, we deal with the problem of 3D mesh generation and representation on a fixed topology. The setting of fixed topology is currently being studied in computer vision and graphics applications and is significantly easier, since it is assumed that the mesh is given and the vertices are canonically ordered; the generation problem thus amounts only to determining the embedding of the mesh.  

The first intrinsic convolutional autoencoder architecture on meshes (MeshVAE) was shown in \cite{litany2017deformable}. The authors used convolutional operators from \cite{verma2018feastnet} and showed examples of human body shape completion from partial scans. 
A follow-up work CoMA~\cite{ranjan2018generating} used a similar architecture with spectral Chebyshev filters~\cite{defferrard2016convolutional} and additional spatial pooling to generate 3D facial meshes. The authors claim that CoMA can represent better faces with expressions than PCA in a very small dimensional latent space of only eight dimensions. In this paper, we present the first GANs structure for generating meshes of 3D faces with fixed topology.

\subsection{Generative adversarial networks}
GANs are a promising unsupervised machine learning methodology implemented by a system of two deep neural networks competing against each other in a zero-sum game framework~\cite{goodfellow2014generative}. GANs has become hugely popular owing to their capability of modeling the distribution of visual data and generating new instances that have many realistic characteristics (\ie, preserving the high-frequency details) and look authentic to human observers. Currently, GANs are among the top choices to generate visual data and they are preferable to auto-encoders and VAEs~\cite{lucic2017gans}. 

Nevertheless, the original GANs were criticized for being difficult to train and prone to mode collapse. Different GANs were proposed to tackle these problems. Wasserstein GAN (WGAN)~\cite{arjovsky2017wasserstein} proposed a new loss function using Wasserstein distance to stabilize the training. In continuation of WGAN, Gulrajani~\etal~\cite{gulrajani2017improved} proposed an alternative way to clip weights, which helped to improve the training convergence and generation quality. 
Boundary Equilibrium GANs (BEGANs)~\cite{berthelot2017began} implemented the discriminator as an auto-encoder whose loss is derived from Wasserstein distance. In that, an equilibrium enforcing method was proposed to balance the training of generator and discriminator. Chang~\etal~\cite{chang2018escaping} further proposed a variant of BEGAN with a Constrained Space (BEGAN-CS). They tried to improve the training stability by adding a latent-space constraint in the loss function. As BEGAN has demonstrated good performance in generating photo-realistic faces, following BEGAN, we meticulously design a generative network for realistic generation of 3D faces. 

\subsection{3D Facial shape representation and generation}
For the past two decades, the method of choice for representing and generating 3D faces is still Principal Component Analysis (PCA). PCA was used for building statistical 3D shape model (\ie, 3D Morphable Models (3DMMs)) in many works~\cite{paysan20093d,patel20093d, blanz1999morphable}. Recently, PCA is adopted for building large scale statistical models of the 3D face~\cite{booth2018large} and head~\cite{Dai_2017_ICCV}. It is very convenient for representing and generating faces to decouple facial identity variations from expression variations. Hence, statistical blendshape models have been introduced which represent only the expression variations using PCA~\cite{li2017learning,neumann2013sparse} or multilinear methods~\cite{brunton2014multilinear,bolkart2015groupwise}. Some recent efforts were made to represent facial expressions with deep learning using fully connected layers~\cite{tan2018variational,tran2018nonlinear}. Fully connected layers have huge number of parameters and also do not take into account the local geometric of the 3D facial surfaces. The only method that represented faces using convolutions on the mesh domain was the recently proposed mesh auto-encoder CoMA~\cite{ranjan2018generating}. Nevertheless, the identity and expression latent space of CoMA was mixed. Furthermore, the representative power and expressiveness of the model is somewhat limited because it was trained on only 12 subjects displaying 12 classes of extreme expressions. In this paper, we train deep generative graph convolutional neural networks (DGCNs) using spectral mesh convolutions that individually model identity and expression on large scale data.

\begin{figure*}[ht!]
    \begin{flushright}
        \begin{minipage}[t]{0.65\linewidth}
          \centering
          \includegraphics[width=\linewidth]{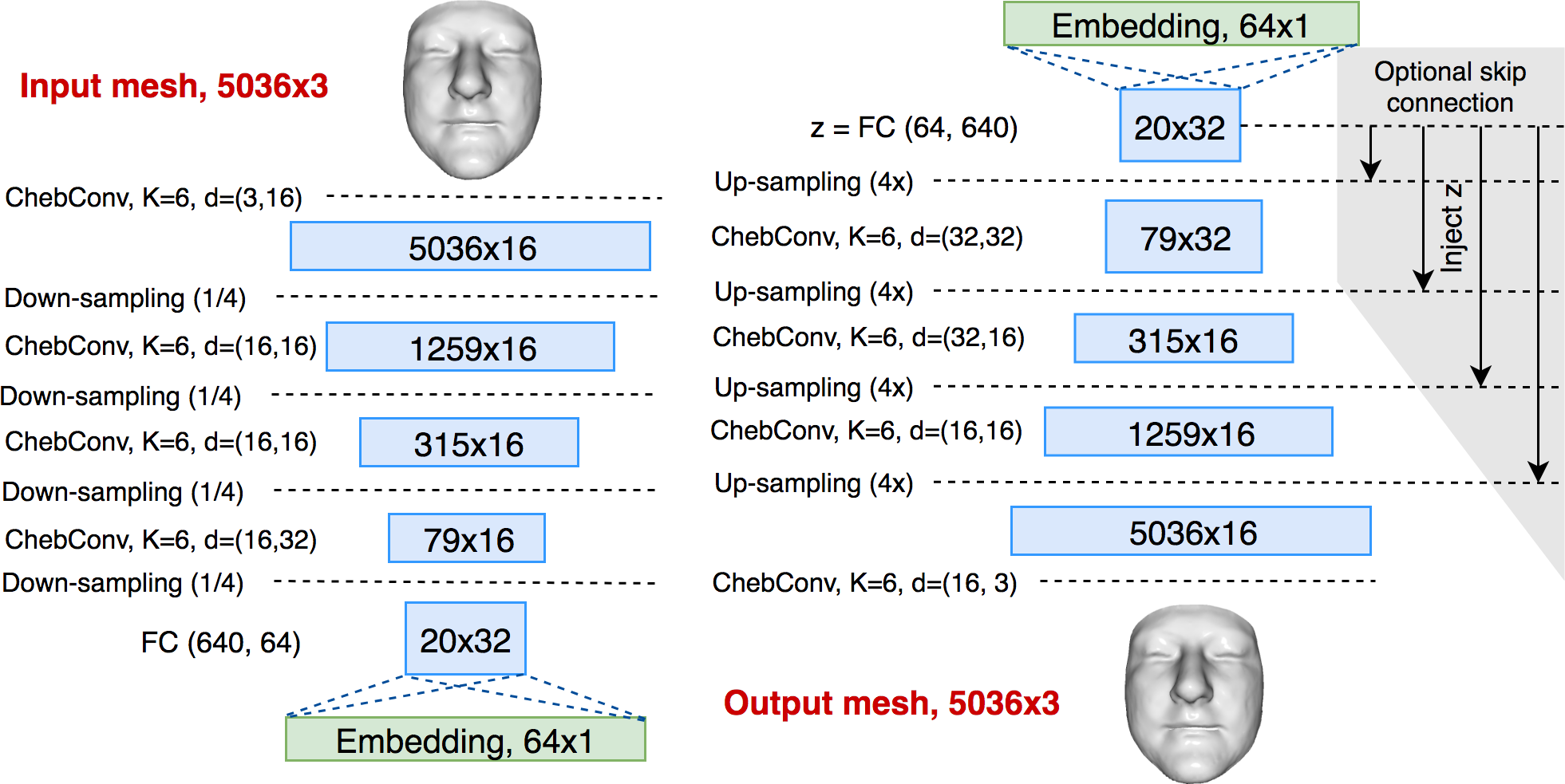}
          \vspace{-5pt}
          \caption{Network architecture of the proposed MeshGAN.}
          \label{fig:meshgan}
        \end{minipage}%
        \hfill
        \begin{minipage}[t]{0.35\linewidth}
          \centering
          \includegraphics[width=0.75\linewidth, height=1.1\linewidth]{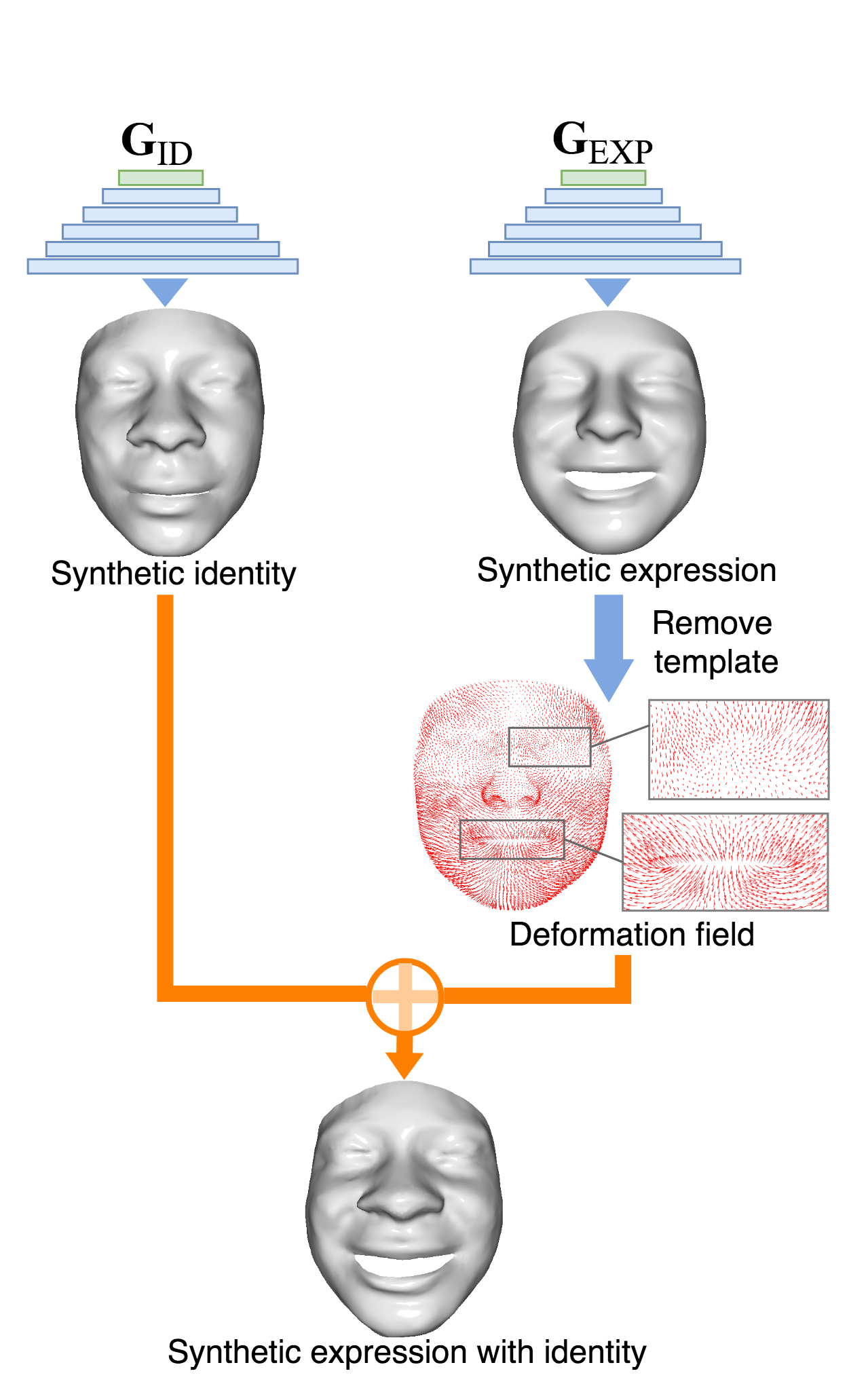}
          \vspace{-5pt}
          \caption{Our pipeline to generate random 3D face with expression.}
          \label{fig:two_gan}
        \end{minipage}%
    \end{flushright}
\end{figure*}
\section{Proposed Approach}
In this part, we define the mesh convolution operators, describe our encoder and decoder/generator and layout our MeshGAN architecture for non-linear generation of 3D faces.

\subsection{Data representation}
We represent the facial surface as a manifold triangular mesh $\mathcal{M} = (\{1, \hdots, n\},\mathcal{E} = \mathcal{E}_i \cup \mathcal{E}_b,\mathcal{F})$ 
where each edge $e_{ij}\in \mathcal{E}$ belongs to at most two triangle faces $F_{ijk}$ and $F_{jih}$ (here, we denote by $\mathcal{E}_i$ and $\mathcal{E}_b$ the interior and boundary edges, respectively).
An {\em embedding} of $\mathcal{M}$ is realised by assigning 3D coordinates to the vertices $V$, which are encoded as a $n\times 3$ matrix $\mathbf{V}$ containing the vertex coordinates as rows. 
%
%
The discrete Riemannian metric is defined by assigning a length $\ell_{ij} = \| \mathbf{v}_i - \mathbf{v}_j \|_2$ to each edge $e_{ij}\in \mathcal{E}$. 

The Laplacian operator is discretised (using the distance-based equivalent of the cotangent formula \cite{jacobson2012cotangent,meyer2003discrete}) as an $n\times n$ matrix $\boldsymbol{\Delta} = \mathbf{A}^{-1} \mathbf{W}$, where $\mathbf{A}$ is a diagonal matrix of local area elements $a_i=\frac{1}{3}\sum_{jk:ijk\in F} a_{ijk}$, and $\mathbf{W}$ is a symmetric matrix of edge-wise weights, defined in terms of the discrete metric:
%
\begin{equation}\label{eq:w}
w_{ij}=
\begin{cases}
\frac{-\ell^2_{ij}+\ell^2_{jk}+\ell^2_{ki}}{8 A_{ijk}} + \frac{-\ell^2_{ij}+\ell^2_{jh}+\ell^2_{hi}}{8 A_{ijh}}, & \mbox{if $e_{ij} \in \mathcal{E}_i$};\\
\frac{-\ell^2_{ij}+\ell^2_{jh}+\ell^2_{hi}}{8 A_{ijh}}, & \mbox{if $e_{ij} \in \mathcal{E}_b$};\\
-\sum_{k\neq i} w_{ik}, & \mbox{if $i=j$}.
\end{cases}
\nonumber
\end{equation}
The Laplacian admits an eigen decomposition 
 $\mathbf{\Delta} = \mathbf{\Phi \Lambda \Phi}^\top$ with $\mathbf{A}$-orthonormal eigenvectors $\boldsymbol{\Phi} = (\boldsymbol{\phi}_1^\top, \hdots, \boldsymbol{\phi}_n^\top)$ and non-negative eigenvalues $0 = \lambda_1 \leq \lambda_2 \leq  \hdots  \lambda_n$ arranged into a diagonal matrix $\boldsymbol{\Lambda} = \mathrm{diag}(\lambda_1, \hdots, \lambda_n)$. 

\subsection{Spectral mesh convolutions}

Let $\mathbf{f} = (f_1, \hdots, f_n)^\top$ be a scalar real function defined on the vertices of the mesh, represented as an $n$-dimensional vector. 
%
The space of such functions is a Hilbert space with the standard inner product $\langle \mathbf{f}, \mathbf{g} \rangle = \mathbf{f}^\top \mathbf{A} \mathbf{g}$. 
The eigenvectors of the Laplacian form an orthonormal basis in the aforementioned Hilbert space, allowing a Fourier decomposition of the form $\mathbf{f}  = \boldsymbol{\Phi} \boldsymbol{\Phi}^\top \mathbf{A}\mathbf{f}$, where $\hat{\mathbf{f}} = \boldsymbol{\Phi}^\top\mathbf{A}\mathbf{f}$ is the {\em Fourier transform} of $\mathbf{f}$. The Laplacian eigenvectors thus play the role of standard Fourier atoms and the corresponding eigenvalues that of the respective frequencies. 
Finally, a convolution operation can be defined in the spectral domain by analogy to the Euclidean case as $\mathbf{f} \star \mathbf{g} =  \mathbf{\Phi} ( \hat{\mathbf{f}}  \cdot \hat{\mathbf{g}} ) = \mathbf{\Phi} (\mathbf{\Phi}^\top\mathbf{A} \mathbf{f}) \cdot (\mathbf{\Phi}^\top \mathbf{A}\mathbf{g})$.

{\bf Spectral graph CNNs.} Bruna \etal \cite{bruna2013spectral} exploited the above formulation for designing graph convolutional neural networks, in which a basic spectral convolution operation has the form 
$\mathbf{f}' =   \boldsymbol{\Phi} \hat{\mathbf{G}} \boldsymbol{\Phi}^\top \mathbf{f}$, 
%
where $\hat{\mathbf{G}} = \mathrm{diag}(\hat{g}_{1}, \hdots, \hat{g}_{n})$ is a diagonal matrix of spectral multipliers representing the filter and $\mathbf{f}'$ is the filter output. 
Among notable drawbacks of this architecture putting it at a clear disadvantage compared to classical Euclidean CNNs are: high computational complexity ($\mathcal{O}(n^2)$ due to the cost of computing the forward and inverse graph Fourier transform, incurring dense $n\times n$ matrix multiplication), $\mathcal{O}(n)$ parameters per layer, and no guarantee of spatial localization of the filters.

%

{\bf ChebNet.} Defferrard \etal \cite{defferrard2016convolutional} considered the spectral CNN framework with polynomial filters represented in the Chebyshev basis, $\tau_{\boldsymbol{\theta}}(\lambda) = \sum_{j=0}^p \theta_j T_j(\lambda)$, where $T_j(\lambda) = 2\lambda T_{j-1}(\lambda) - T_{j-2}(\lambda)$ denotes the Chebyshev polynomial of degree $j$, with $T_1(\lambda) =\lambda$ and $T_0(\lambda) =1$. 
A single filter of this form can be efficiently computed by applying powers of the  Laplacian to the feature vector, 
\begin{equation} \label{eq:filt_cheby}
	{\mathbf{f}}' = \mathbf{\Phi} \sum_{j=0}^{p} \theta_j T_j(\tilde{\boldsymbol{\Lambda}}) \mathbf{\Phi}^\top \mathbf{A}\mathbf{f} = \sum_{j=0}^{p} \theta_j T_j(\mathbf{\tilde{\Delta}})  \mathbf{f},
\end{equation}
thus avoiding its eigendecomposition altogether. Here $\tilde{\lambda}$ is a frequency rescaled in $[-1,1]$, $\tilde{\boldsymbol{\Delta}} = 2 \lambda_{n}^{-1}\boldsymbol{\Delta}  - \mathbf{I}$ is the rescaled Laplacian with eigenvalues $\tilde{\boldsymbol{\Lambda}} = 2 \lambda_{n}^{-1} \boldsymbol{\Lambda}  - \mathbf{I}$. The computational complexity thus drops from $\mathcal{O}(n^2)$ as in the case of spectral CNNs to $\mathcal{O}(n)$, since the mesh is sparsely connected. 
%


\subsection{MeshGAN}
We introduce MeshGAN, a variant of BEGAN~\cite{berthelot2017began}, that can learn a non-linear 3DMM directly from the 3D meshes. Specifically, we employ the aforementioned ChebNet to build our discriminator $D$ and generator $G$.

\subsubsection{Boundary equilibrium generative adversarial networks}
The main difference between BEGAN and typical GANs is that, BEGAN uses an auto-encoder as the discriminator, as it tries to match the auto-encoder loss distribution rather than the data distributions. This is achieved by adding an extra equilibrium term $\gamma\in[0,1]$. More precisely, this hyper-parameter is used to maintain the balance of the loss expectation of discriminator $D$ and generator $G$ (\ie, $\mathbb{E}[\mathcal{L}(G(\mathbf{z}))] = \gamma\mathbb{E}[\mathcal{L}(\mathbf{x})]$). The training objective of BEGAN is as follows:
\begin{equation}
    \left\{ \begin{array}{ll}
    \mathcal{L}_D = \mathcal{L}(\mathbf{x}) - k_t \cdot \mathcal{L}(G(\mathbf{z}_D)), \,\, & \text{for } \mathbf{\theta}_D; \\ 
    \mathcal{L}_G = \mathcal{L}(G(\mathbf{z}_G)), \,\, & \text{for } \mathbf{\theta}_G; \\
    k_{t+1} = k_{t} + \lambda_k (\gamma \mathcal{L} - \mathcal{L}(G(\mathbf{z}_G))), \,\, & \text{in train step t}, \nonumber \\
    \end{array} \right.
\end{equation}
where $\mathbf{z}\in [{-1}, 1]^{h}$ is the uniform random vector of dimension $h$ (aka. the latent vector of generator), $\mathbf{\theta}_D$ and $\mathbf{\theta}_G$ are the trainable parameters of the discriminator and generator respectively; $\mathcal{L}(\cdot)$ is the discriminator loss, for which we select $L1$ loss in this paper. In each training step $t$, variable $k_t \in [0,1]$ is utilised to control the influence of the fake loss $\mathcal{L}(G(\mathbf{z}))$ on discriminator; $\lambda_k$ can be regarded as the learning rate of $k$, which is set to 0.001. Berthelot~\etal~\cite{berthelot2017began} found out that $\gamma$ has a decisive impact on the diversity of generated images, that is, lower values tends to produce mean face-alike images. To encourage more variations, we empirically set $\gamma$ to 0.7.

\subsubsection{MeshGAN architecture}
Based on the architecture of BEGAN, we developed MeshGAN using ChebNet~\cite{defferrard2016convolutional,kipf}. The architecture of MeshGAN is illustrated in Fig.~\ref{fig:meshgan}. We follow a similar design of CoMA for building our encoder and generator/decoder, 4 Chebyshev convolutional filters with $K$ = 6 polynomials are used in the encoder. Nevertheless, after each convolution layer, we select ELU~\cite{clevert2015fast} as the activation function to allow the passing of negative values. The mesh down-sampling step is performed by the surface simplification method in~\cite{garland1997surface}, which minimises the quadric error when decimating the template. Up-sampling of the template is based on the barycentric coordinates of the contracted vertices in decimated mesh~\cite{ranjan2018generating}. In total, we perform 4 levels of down-sampling, with each level lowering the number of vertices by approximately 4 times. To allow for more representation powers, we set the bottleneck of discriminator to be 64, equal to the dimension of feature embedding in generator. Momentum optimizer~\cite{qian1999momentum} is employed, with the learning rate being $0.008$ and decay rate being $0.99$. We train all the models with 300 epochs. Note that skip connections between the output of fully connected layer and each up-sampled graph can be applied to encourage more facial details.

\section{Experiments}

\subsection{3D face databases}
\noindent \textbf{3dMD:} For identity model training, we used recently collected 3dMD datasets scanned by the high resolution 3dMD device\footnote{http://www.3dmd.com/}. We selected around 12,000 unique identities from this database, with different ethnic groups (\ie, Chinese, Caucasian, Black people) and age groups presented. 

\noindent \textbf{4DFAB:} To train expression models, we use the 4DFAB database~\cite{cheng20184dfab}, which is the largest dynamic 3D face database that contains both posed and spontaneous expressions. In 4DFAB, participants were invited to attend four experiment sessions at different times. In each session, participants were asked to articulate 6 basic facial expressions, and then watched several emotional videos. Annotation of apex posed expression frames as well as the expression category of spontaneous sequences were provided. To ensure the richness of expressions in our training set, we randomly sampled 6,651 apex posed expression meshes and 7,567 spontaneous expression meshes from 4DFAB.

For each database, we train the CoMA and MeshGAN model with the corresponding data. We label the models that are trained on 3dMD database with \textbf{-ID}, whereas the models trained on 4DFAB database are appended with \textbf{-EXP}.

\subsubsection{Data pre-processing}
To balance the fineness and complexity of model, we cropped and decimated the LSFM model~\cite{booth2018large}, and generated a new 3D template with 5,036 vertices. In order to bring all the data into dense correspondence with the template, we employed Non-rigid ICP~\cite{amberg2007optimal} to register each mesh. We automatically detected 79 3D facial landmarks with the UV-based alignment method developed in~\cite{cheng20184dfab}, and utilised these landmarks to assist dense registration. Unless otherwise stated, we divided each database into training and testing sets with a split ratio of 9:1.

On a separate note, in order to train the expression models, we need to decouple facial identity from every expression mesh in 4DFAB. This was achieved by manually selecting one neutral face per subject per session in 4DFAB, and subtracting the expression mesh with its corresponding neutral face to obtain the facial deformation. We then exerted this deformation on the 3D template to generate a training set with pure expressions. Note that a local surface-preserving smoothing step~\cite{taubin1996optimal} was undertaken to further remove identity information as well as noises.
\begin{figure*}[ht!]
\centering
    \includegraphics[width=0.95\linewidth]{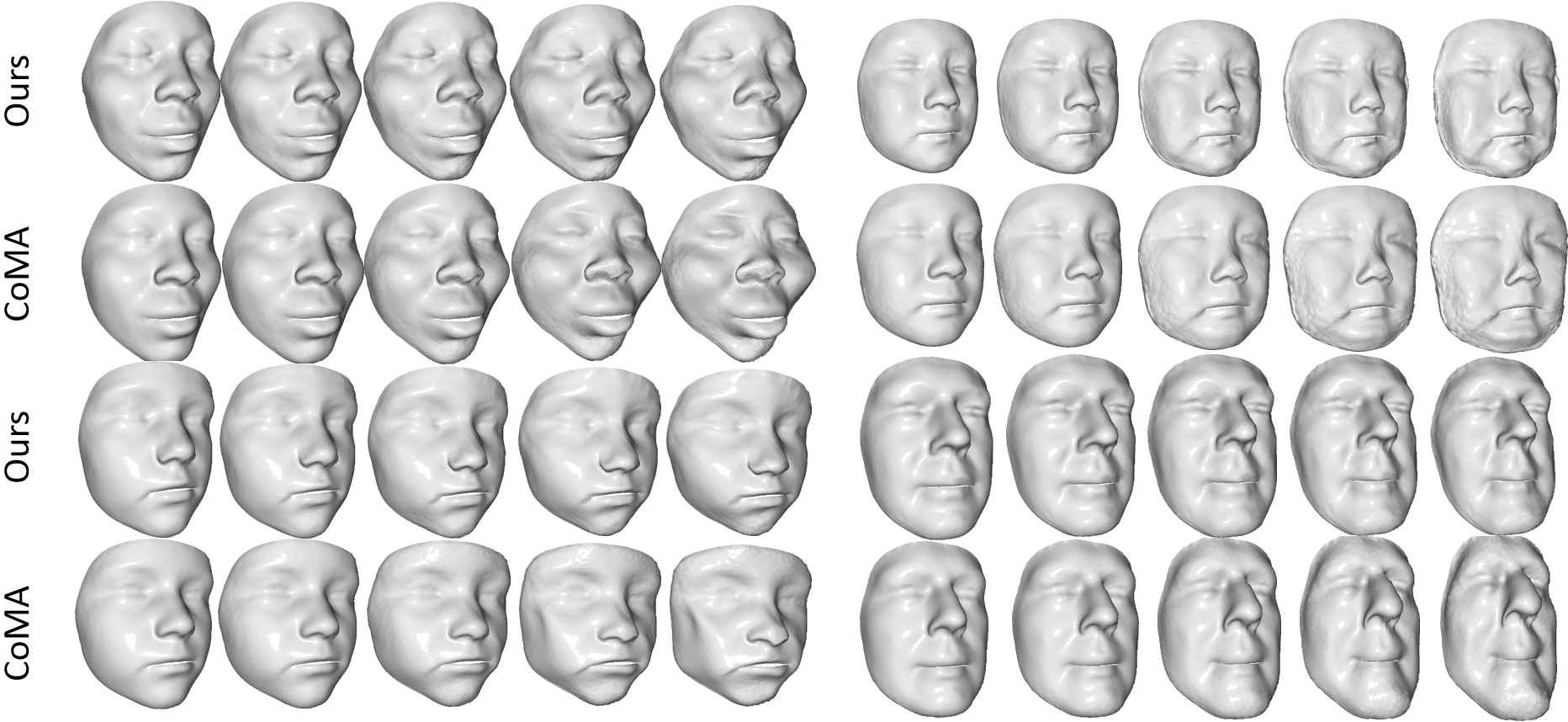}
    \vspace{-2pt}
\caption{Extrapolation of the identity model. First 2 rows are examples of exaggerating ethnicity (Black people and Chinese). Last 2 rows display exaggerated ages (children and elderly people). Please check our supplementary material for extrapolation result of gender.}
\label{fig:extrap_id}
\end{figure*}
\begin{figure*}[ht!]
\centering
    \includegraphics[width=0.95\linewidth]{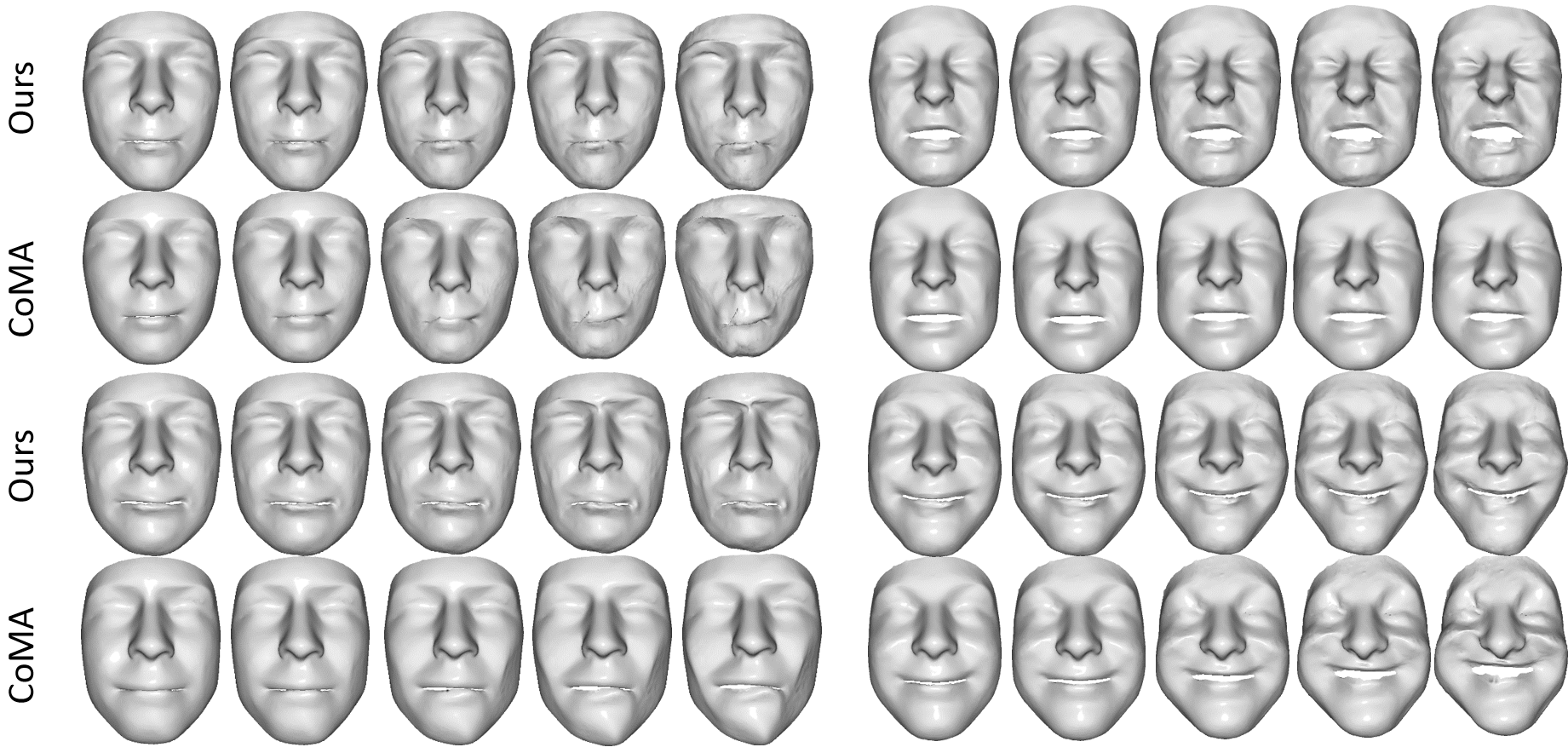}
    \vspace{-2pt}
\caption{Extrapolation of the expression model. First 2 rows are examples of exaggerating anger and disgust. Last 2 rows are extrapolation of sad and happy. Please check our supplementary material for other expressions.}
\label{fig:extrap_expr}
\end{figure*}
\begin{table}[]
    \centering
    \begin{tabular}{@{}cccc@{}}
    \toprule
    Methods & Generalisation & Specificity & FID \\
    \midrule
    CoMA-ID & \textbf{0.442$\pm$0.116} & 1.60$\pm$0.228 & 14.24 \\
    MeshGAN-ID  & 0.465$\pm$0.189 & \textbf{1.433$\pm$0.144} & \textbf{10.82}\\
    \hline 
    \end{tabular}
    \\
    \vspace{-5pt}
    \caption{Intrinsic evaluation of identity models. Average generalisation and specificity errors are measured in mm.}
    \label{tab:eval_id}
\end{table}
\begin{table}[]
    \centering
    \begin{tabular}{@{}cccc@{}}
    \toprule
    Methods & Generalisation & Specificity & FID \\
    \midrule
    CoMA-EXP       & 0.606$\pm$0.203 & 1.899$\pm$0.272 & 22.43\\
    MeshGAN-EXP & \textbf{0.605$\pm$0.264} & \textbf{1.536$\pm$0.153} & \textbf{13.59}\\
    \hline 
    \end{tabular}
    \\
    \vspace{-5pt}
    \caption{Intrinsic evaluation of expression models. Average generalisation and specificity errors are measured in mm.}
    \label{tab:eval_expr}
\end{table}
\begin{figure}[htb!]
    \centering
    \includegraphics[width=0.9\linewidth]{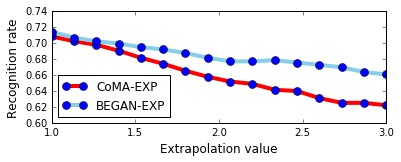}\\
    \vspace{-10pt}
    \caption{3D facial expression recognition on exaggerated expressions generated by extrapolating latent space of CoMA and MeshGAN.}
    \label{fig:3dfer}
\end{figure}
\subsection{Intrinsic evaluation of MeshGAN} 
We gave a quantitative evaluation of MeshGAN's generator, whose counterpart is the decoder of CoMA. The intrinsic characteristics of the models include generalisation capability, specificity~\cite{brunton2014review, bolkart2015groupwise}, as well as FID score~\cite{heusel2017gans}.

\noindent \textbf{Generalisation.} The generalisation measures the ability of a model to represent/reconstruct unseen face shapes that are not present during training. To compute the \emph{generalisation error}, we computed the per-vertex Euclidean distance between every sample $\mathbf{x}\in \mathrm{R}^{n\times3}$ of the test set and its corresponding reconstruction $\mathbf{x}^{*}$ by the generator $G(\mathbf{z}), \, \mathbf{z}\in\mathrm{R}^{h\times1}$:
\begin{equation}
    \mathbf{x}^{*} = \underset{\mathbf{z}}{\mathrm{argmin}} | \mathbf{x} - G(\mathbf{z})|.
    \label{eq:optim_z}
\end{equation}
After that, we took the average value over all vertices and all test samples. This procedure was conducted separately on identity and expression models. We reported the mean and standard deviation of the reconstruction errors in Table~\ref{tab:eval_id} and Table~\ref{tab:eval_expr}. It can be seen that both methods achieved similar performance in reconstructing facial expressions (MeshGAN-ID achieved 0.605mm, while CoMA-ID produced 0.606mm), whereas CoMA is slightly better in describing unseen identity (0.023mm lower in error). This is probably attributed to the fact that auto-encoder is specifically trained to reconstruction data examples, while BEGAN is not. We leave this as our future investigation, and refer the readers to~\cite{creswell2018inverting, zhu2016generative}.

\noindent \textbf{Specificity.} The specificity of a model evaluates the validity of generated faces. For each model, we randomly synthesised 10,000 faces and measured the proximity between them and the real faces in test set. More precisely, for every randomly generated face, we found its nearest neighbor in the test set, in terms of minimum (over all samples of the test set) of the average per-vertex distance. We recorded the mean and standard deviation of this distance over all random samples as the \emph{specificity error}. Note that we randomly sampled MeshGAN with the uniform distribution $\mathcal{U}(-1,1)$, whereas we facilitated CoMA with a multivariate Gaussian distribution $\mathcal{N}(\mathbf{\mu}_c, \mathbf{\Sigma}_c)$ estimated from the features embedding of the training data in CoMA (using Eq.~\ref{eq:optim_z}). Table~\ref{tab:eval_id} and Table~\ref{tab:eval_expr} also display the specificity errors for different models. We observed that in all the cases, MeshGAN attained particularly low errors against CoMA, \ie, 0.17mm lower in identity, 0.36mm lower in expression. This is a quantitative evidence that the synthetic faces generated by MeshGAN models are more realistic than those of CoMA.

\noindent \textbf{Fr{\'e}chet Inception Distance (FID).} FID~\cite{heusel2017gans} is a reliable measurement on the quality and diversity of the images generated by GANs. To compute \emph{FID score}, we borrowed the pre-trained Inception network~\cite{szegedy2015going} to extract features from an intermediate layer and then modelled the distribution of these features using a multivariate Gaussian $\mathcal{N}(\mathbf{\mu}, \mathbf{\Sigma})$. As Inception network is trained on 2D images, we rasterised each 3D mesh (with lambertian shading) into a 64$\times$64 image and fed it to the network. The FID score between the real images $\mathcal{I}_R$ and generated images $\mathcal{I}_G$ is computed as:
\begin{equation}
    \text{FID}(\mathcal{I}_R, \mathcal{I}_G) = \| \mathbf{\mu}_R - \mathbf{\mu}_G \|^2_2 + Tr(\mathbf{\Sigma}_R + \mathbf{\Sigma}_G - 2\sqrt{\mathbf{\Sigma}_R \mathbf{\Sigma}_G})), \nonumber
\end{equation}
where $\mathcal{N}(\mathbf{\mu}_R, \mathbf{\Sigma}_R)$ and $\mathcal{N}(\mathbf{\mu}_G, \mathbf{\Sigma}_G)$ are the multivariate Gaussians estimated from the inception feature of the real and generated images respectively. The smaller the FID values are, the better the image quality and diversity would be. It has to be mentioned that, when sampling the latent space of CoMA, we did not estimate the multivariate Gaussian beforehand, as the training data distribution is not supposed to be revealed here. Hence, we used a standard Gaussian $\mathcal{N}(0,1)$ to sample latent space of CoMA, meanwhile for the MeshGAN, we always use the uniform distribution $\mathcal{U}(-1,1)$. We show the FID scores of CoMA and MeshGAN in Table~\ref{tab:eval_id} and Table~\ref{tab:eval_expr}. We can observe that FID scores of MeshGAN are significantly lower than those of CoMA in both cases. This is another strong evidence that MeshGAN can generate meshes with richer variations and better quality than auto-encoders. 

As a matter of fact, we also experimented with different GANs (such as the original GANs~\cite{goodfellow2014generative}, WGAN~\cite{arjovsky2017wasserstein} and BEGAN-CS~\cite{chang2018escaping}) in the same architectures as MeshGAN. Unfortunately, they did not achieve any comparable performances with CoMA or BEGAN. Due to limited space, we put this ablation study in the supplementary material.
\begin{figure*}[ht!]
    \centering
    \includegraphics[width=0.9\linewidth,height=0.95\linewidth]{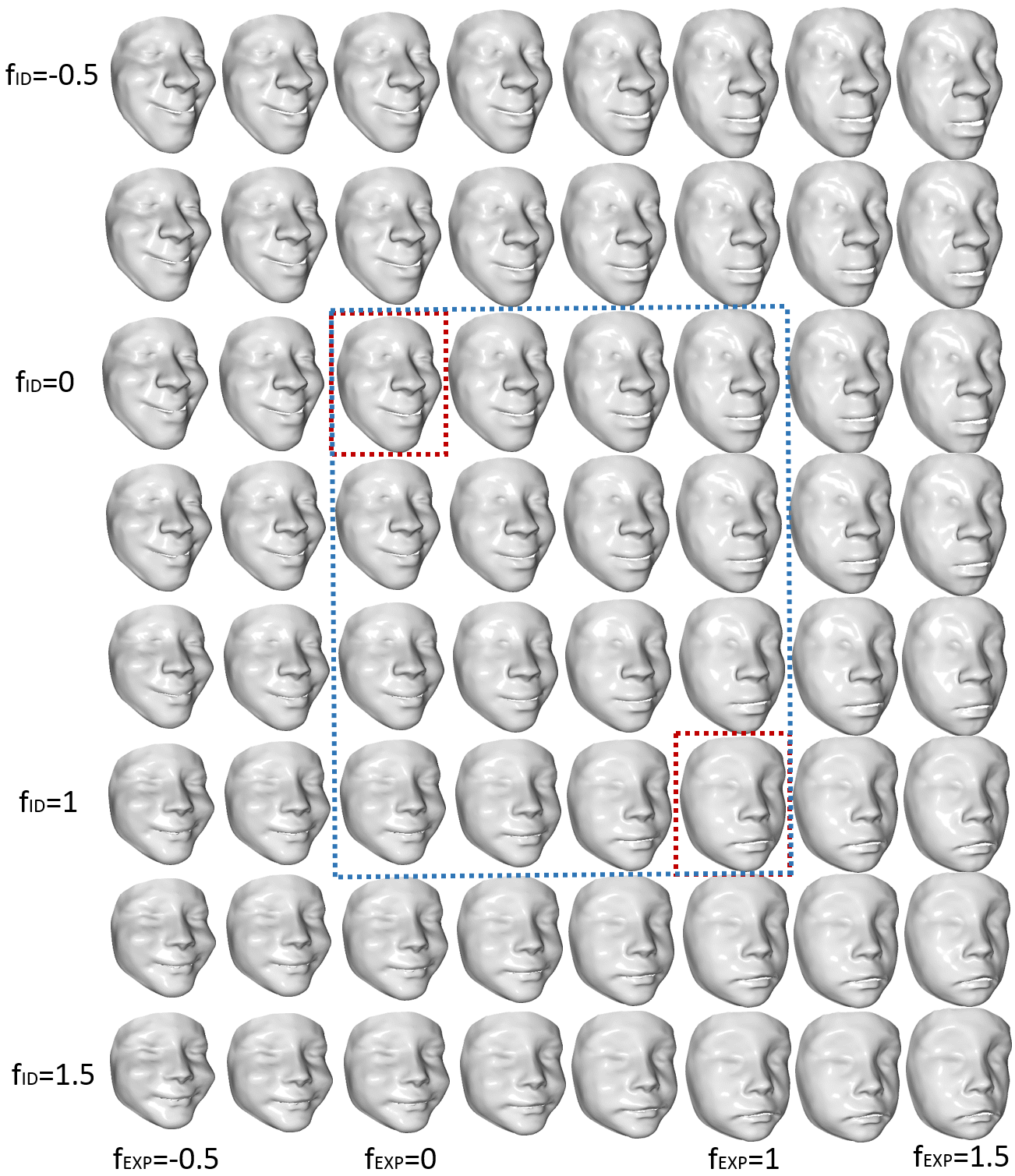}
    \vspace{-10pt}
    \caption{Interpolation and extrapolation between two generated meshes (\ie, the anchor meshes marked in red box). The blue box contains interpolation results between two anchor meshes, the outside faces are extrapolation results.}
    \label{fig:interp}
\end{figure*}
\subsection{Extrapolating identity and expression model}
We first extrapolated the latent vector of the identity model and visualised the exaggerated synthetic examples. Given a pair of meshes $\mathbf{x}_1$ and $\mathbf{x}_2$, we estimated the feature embedding (denoted as $\mathbf{z}$) using Eq.~\ref{eq:optim_z}. After that, we computed the extrapolated latent vector using a non-convex combination of two vectors $\mathbf{z}_1$ and $\mathbf{z}_2$: 
\begin{equation}
    \tilde{\mathbf{z}} = (1-f)\mathbf{z}_1 + f\mathbf{z}_2, \,\, \text{when } f < 1 \,\, \text{or } \,\, f > 1.
    \label{eq:nonconvex_comb}
\end{equation}
Here, we fixed mesh $\mathbf{x}_1$ to be the neutral template, while $\mathbf{x}_2$ was the target face reconstructed by MeshGAN and CoMA, separately. Fig.~\ref{fig:extrap_id} shows the extrapolation results of the identity model in terms of ethnicity and age (note that we increased $f$ from $1$ to $2$). We can clearly observe that: (a) MeshGAN can better describe the subtle facial details (\eg, eyes and lips); (b) CoMA produces highly distorted and grotesque faces (\eg, disproportionate nose, incorrect exaggeration of ethnicity and age) as the extrapolation proceeds, whereas MeshGAN did not have such issues.

For the extrapolation of expression models, we followed the same approach and showed the results in Fig.~\ref{fig:extrap_expr}. Obviously, MeshGAN is more capable of representing different facial expressions, especially the facial muscle movement (\eg, disgust in the first row). Compared with CoMA, the exaggerated expressions from MeshGAN are still quite meaningful and realistic. To quantitatively evaluate the semantic correctness of exaggerated expressions, we trained a 3D expression classifiers using SplineCNN~\cite{fey2018splinecnn}. We built this FER network with 4 convolution layers: SConv($\mathbf{k}$,1,16)$\rightarrow$Pool(4)$\rightarrow$SConv($\mathbf{k}$,16,16)$\rightarrow$Pool(4)$\rightarrow$\\
SConv($\mathbf{k}$,16,16)$\rightarrow$Pool(4)$\rightarrow$SConv($\mathbf{k}$,16,32)$\rightarrow$Pool(4)$\rightarrow$\\
FC(6), where $\mathbf{k} = \{k_1, k_2, k_3\}, k_1=k_2=k_3=5$ are the B-spline kernel sizes. ELU~\cite{clevert2015fast} is used after each convolution and fully connected layer. We trained the network with 80 epochs, learning rate and epoch size equal to 0.0001 and 16, respectively. The Pool($\cdot$) operation is exactly the same as MeshGAN. For FER training, we prepared around 6k posed expression meshes (6 expressions, each has nearly 1k samples) from 4DFAB, which are not present in training set of expression model. We testified the exaggerated expressions produced by different extrapolating factor $f$ (ranged from 1 to 3). We plotted the recognition rate for each $f$ as a curve in Fig.~\ref{fig:3dfer}. Interestingly, as the degree of extrapolation increases, the recognition rate for CoMA drastically declines, while MeshGAN decreases comparatively slowly. This further proves that MeshGAN can still provide meaningful expressions even when sampling beyond the normal range.
%

\subsection{Qualitative results}
We used the pipeline in Fig.~\ref{fig:two_gan} to generate 3D identities with expressions. Qualitative results are shown in Fig.~\ref{fig:intro} (b). To visualise the interpolation and extrapolation between/beyond two faces, we synthesised two identities with different expression and used them as the anchor faces. Following Eq.~\ref{eq:nonconvex_comb}, we varied the parameters of identity and expression models by separate factors $\text{f}_{\text{ID}}$ and $\text{f}_{\text{EXP}}$. By using the grid of interpolated/extrapolated parameters, we synthesised the corresponding faces and displayed them in Fig.~\ref{fig:interp}. 

\section{Conclusion} 

We presented the first GANs capable of generating 3D facial meshes of different identities and different expressions. We have experimentally and empirically demonstrated that the proposed MeshGAN can generate 3D facial meshes with more subtle details than the state-of-the-art auto-encoders. Finally, we show that the proposed MeshGAN can model the distribution of faces better than auto-encoders, hence it leads to better sampling strategies.

{\small
\bibliographystyle{ieee}
\bibliography{egpaper_for_review}
}

\end{document}